\documentclass{ecai} 

\usepackage{latexsym}
\usepackage{amssymb}
\usepackage{amsmath}
\usepackage{amsthm}
\usepackage{booktabs}
\usepackage{enumitem}
\usepackage{graphicx}
\usepackage{color}
\usepackage{thmtools}
\usepackage{thm-restate}

\usepackage[utf8]{inputenc} 
\usepackage[T1]{fontenc}    
\usepackage[colorlinks]{hyperref}       
\usepackage{url}            
\usepackage{booktabs}       
\usepackage{amsfonts}       
\usepackage{nicefrac}       
\usepackage{microtype}      
\usepackage{xcolor}         
\usepackage{mathtools}
\usepackage{algorithm}
\usepackage[noend]{algpseudocode}
\usepackage{todonotes}
\usepackage{amsthm}
\usepackage{adjustbox}
\usepackage{todonotes}
\usepackage{amssymb}
\usepackage{comment}
\usepackage{wrapfig}
\usepackage{subcaption}

\newtheorem{theorem}{Theorem}
\newtheorem{lemma}[theorem]{Lemma}
\newtheorem{corollary}[theorem]{Corollary}

\newtheorem{definition}{Definition}
\newtheorem{assumption}[theorem]{Assumption}

\newcommand{\BibTeX}{B\kern-.05em{\sc i\kern-.025em b}\kern-.08em\TeX}

\newcommand{\tSi}{\tau}   

\newcommand{\cC}{\mathcal{C}}

\newcommand{\cE}{\mathcal{E}}
\newcommand{\cJ}{\mathcal{J}}
\newcommand{\cL}{\mathcal{L}}

\newcommand{\cP}{\mathcal{P}}
\newcommand{\cR}{\mathcal{R}}
\newcommand{\cS}{\mathcal{S}}
\newcommand{\cT}{\mathcal{T}}

\newcommand{\cX}{\mathcal{X}}
\usepackage{bbm, dsfont}
\newcommand{\allones}{\mathbbm{1}}
\newcommand{\indicator}[1]{\mathbb{I}\{#1\}}

\newcommand{\EEc}[2]{\mathbb{E}\left[#1\;\middle\lvert\;#2\right]}

\newcommand{\pa}[1]{\left(#1\right)}

\newcommand{\norm}[1]{\left\|#1\right\|}

\newcommand{\infnorm}[1]{\norm{#1}_\infty}

\newcommand{\real}{\mathbb{R}}
\definecolor{mygreen}{HTML}{569e34}

\hypersetup{
    citecolor=blue,
    menucolor=.,
}

\begin{document}


\begin{frontmatter}


\paperid{2075} 


\title{Hierarchical Average-Reward\\
Linearly-solvable Markov Decision Processes}


\author[A]{\fnms{Guillermo}~\snm{Infante}\thanks{Corresponding Author. Email: guillermo.infante@upf.edu}}
\author[A]{\fnms{Anders}~\snm{Jonsson}}
\author[A]{\fnms{Vicen\c{c}}~\snm{G\'omez}} 

\address[A]{AI\&ML research group, Universitat Pompeu Fabra, Barcelona, Spain}


\begin{abstract}
We introduce a novel approach to hierarchical reinforcement learning for Linearly-solvable Markov Decision Processes (LMDPs) in the infinite-horizon average-reward setting. Unlike previous work, our approach allows learning low-level and
high-level tasks simultaneously, without imposing limiting restrictions on the low-level tasks. Our method relies on partitions of the state space that create smaller
subtasks that are easier to solve, and the equivalence between such partitions to
learn more efficiently. We then exploit the compositionality of low-level tasks to
exactly represent the value function of the high-level task. Experiments show that
our approach can outperform flat average-reward reinforcement learning by one or
several orders of magnitude.
\end{abstract}

\end{frontmatter}


\section{Introduction}

Hierarchical reinforcement learning (RL)~\citep{Dayan1992,Parr1997,Sutton1999,Dietterich2000,Barto2003} aims to decompose a complex high-level task into several low-level tasks. After solving the low-level tasks, their solutions can be combined to form a solution to the high-level task. Ideally, the low-level tasks should be significantly easier to solve than the high-level task, in which case one can obtain an important speedup in learning~\citep{Nachum2018,Wen2020}. Hierarchical RL has also been credited with other advantages, e.g.~the ability to explore more efficiently~\citep{Nachum2019}.

Most previous work on hierarchical RL considers either the finite-horizon setting or the infinite-horizon setting with discounted rewards.
The average-reward setting is better suited for cyclical tasks characterized by continuous experience.
In the few works on hierarchical RL in the average-reward setting, either the low-level tasks are assumed to be solved beforehand~\citep{Fruit2017,Fruit2017b,Wan2021a} or they 
have important restrictions that severely reduce their applicability, e.g.~a single initial state~\citep{Ghavamzadeh2007}. It is therefore an open question how to develop algorithms for hierarchical RL in the average-reward setting
in order to learn the low-level and high-level tasks simultaneously.

In this paper we propose a novel framework for hierarchical RL in the average-reward setting that simultaneously solves low-level and high-level tasks. Concretely, we consider the class of Linearly-solvable Markov Decision Processes (LMDPs)~\citep{Todorov2006}. 
LMDPs are a class of restricted MDPs for which the Bellman equation can be exactly transformed into a linear equation. This class of problems plays a key role in the framework of RL as probabilistic inference~\cite{Kappen2012,Levine2018}.
One of the properties of LMDPs is compositionality: one can compute the solution to a novel task from the solutions to previously solved tasks without learning~\citep{Todorov2009a}. 
Such a property has been exploited in recent works about compositionality in RL~\cite{Hunt2019,Niekerk2019,NangueTasse2020} and in combination with Hierarchical RL in the finite-horizon setting~\cite{Jonsson2016,Saxe2017,Infante2022}. Adapting this idea to the average-reward setting requires careful analysis.


Unlike most frameworks for hierarchical RL, our proposed approach does not decompose the policy, only the value function. Hence the agent never chooses a subtask to solve, and instead uses the subtasks to compose the value function of the high-level task. 
This avoids introducing non-stationarity at the higher level when updating the low-level policies.

Our work makes the following novel contributions:
\begin{itemize}
  \item Learning low-level and high-level tasks simultaneously in the average-reward setting, without imposing additional restrictions on the low-level tasks.
    \item Two novel algorithms for solving hierarchical RL in the average reward setting: the first one is based on the eigenvector approach used for solving LMDPs. The second is an online variant in which an agent learns simultaneously the low-level and high-level tasks.
    \item Two main theoretical contributions LMDPs: a converge proofs for both differential soft TD-learning for (non-hierarchical) LMDPs and also for the eigenvector approach in the hierarchical case.
\end{itemize}

To the best of our knowledge, this work is the first that extends the combination of compositionality and hierarchical RL to the average-reward setting.

\section{Related work}
Most research on hierarchical RL formulates problems as a Semi-Markov Decision Process (SMDP) with options~\citep{Sutton1999} or the MAXQ decomposition~\citep{Dietterich2000}.

\citet{Fruit2017} and~\cite{Fruit2017b} propose algorithms for solving SMDPs with options in the average-reward setting, proving that the regret of their algorithms is polynomial in the size of the SMDP components, which may be smaller than the components of the underlying Markov Decision Process (MDP).
\citet{Wan2021a} present a version of differential Q-learning for SMDPs with options in the average-reward setting, proving that differential Q-learning converges to the optimal policy.
However, the above work assumes that the option policies are given prior to learning.
\citet{Ghavamzadeh2007} propose a framework for hierarchical average-reward RL based on the MAXQ decomposition, in which low-level tasks are also modeled as average-reward decision processes.
However, since the distribution over initial states can change as the high-level policy changes, the authors restrict low-level tasks to have a single initial state.

\citet{Wen2020} present an approach for partitioning the state space and exploiting the equivalence of low-level tasks, similar to our work. The authors present an algorithm whose regret is smaller than that of algorithms for solving the high-level task.
However, their algorithm does not decompose the high-level task into low-level tasks, but instead exploits the fact that equivalent subtasks have shared dynamics. 
\citet{Infante2022} combine the compositionality of LMDPs with the equivalence of low-level tasks to develop a framework for hierarchical RL in the finite-horizon setting.
Compositionality is also a key idea in the option keyboard~\citep{Barreto2019}, though in the case of MDPs, compositionality is not exact, so the resulting policy will only be approximately optimal.
In contrast, our framework based on LMDPs can represent the globally optimal policy.

\section{Background}

Given a finite set $\cX$, we use $\Delta(\cX)$ to denote the set of all probability distributions on $\cX$.

\subsection{First-exit Linearly-solvable Markov Decision Processes}
A first-exit Linearly-solvable Markov Decision Process (LMDP,~\citet{Todorov2006}) is a tuple
${\cL = \langle\cS,\cT,\cP,\cR,\cJ\rangle}$, where $\cS$ is a set of non-terminal states, $\cT$ is a set of terminal states, ${\cP:\cS\rightarrow\Delta(\cS^+)}$ is the passive dynamics, which describes state transitions in the absence of controls,
${\cR:\cS\rightarrow\real}$ is a reward function on non-terminal states, and ${\cJ:\cT\rightarrow\real}$ is a reward function on terminal states. We use ${S^+=\cS\cup\cT}$ to denote the full set of states.
An agent interacts with the environment following a policy $\pi:\cS\rightarrow\Delta(\cS^+)$.
At \textit{timestep} $t$, it observes a state $s_t$, transitions to a new state $s_{t+1}\sim\pi(\cdot\lvert s_t)$ and receives a reward
\[\cR(s_t, s_{t+1}, \pi) = r_t - \frac 1 \eta\log{\frac{\pi(s_{t+1} \lvert s_t)}{\cP(s_{t+1} \lvert s_t)}},\]
where $r_t$ is a reward with mean $\cR(s_t)$.
The agent can modify the policy $\pi(\cdot\lvert s_t)$, but gets penalized for deviating from the passive dynamics $\cP(\cdot\lvert s_t)$. The parameter $\eta>0$ controls this penalty.
Given $\eta>0$, the value function of a policy $\pi$ can be defined as follows:
\begin{equation}
  v_\eta^\pi(s) = \EEc{\sum_{t=0}^{T-1} \cR(S_t, S_{t+1}, \pi) + \cJ(S_T)}{\pi, S_0 = s},
  \label{eq:value_function_lmdp}
\end{equation}
where $T$ is a random variable representing the length of the episode, and $S_t$, $t\geq 0$, are random variables denoting the state at time $t$. The interaction ends when the agent reaches a terminal state $S_T$ and receives reward $\cJ(S_{T})$. The value function of a terminal state $\tau\in\cT$ is simply
$v_\eta^\pi(\tau) = \cJ(\tau)$. 

The aim of the agent is to find a policy that maximizes expected future reward. For that it is useful to define the \textit{optimal} value function $v_\eta^*:\cS\rightarrow\real$ among all policies. For simplicity, in what follows we omit the subscript and asterisk 
and refer to the optimal value function simply as the value function $v$. Such a value function is known to satisfy the following Bellman
optimality equations~\citep{Todorov2006}:
\begin{equation}
  v(s) = \frac 1 \eta \log{\sum_{s'\in\cS^+}\cP(s'\lvert s)\exp(\eta(\cR(s) + v(s')))} \;\;\forall s\in\cS.
  \label{eq:boe_lmdp}
\end{equation}
Introducing the notation $z(s)= e^{\eta v(s)}$, $s\in\cS^+$, results in the following
system of linear equations:
\begin{equation}
  z(s) = e^{\eta\cR(s)} \sum_{s'\in\cS^+}\cP(s'\lvert s)z(s') \;\;\forall s\in\cS.
  \label{eq:boe_z_lmdp}
\end{equation}
We abuse the notation and for simplicity refer to $z(s)$ and $v(s)$ interchangeably as the value of $s$. Given $z$, an optimal policy is given by
\begin{equation}
  \pi(s'\lvert s) = \frac{\cP(s'\lvert s) z(s')}{ \sum_{s''} \cP(s''\lvert s)z(s'')} \equiv \frac{\cP(s'\lvert s)z(s')}{G[z](s)}.
  \label{eq:optimal_policy}
\end{equation}
The system of linear equations in~\eqref{eq:boe_z_lmdp} can then be written in matrix form when we know the passive dynamics and the reward functions. We let $P \in \real^{\lvert\cS\rvert\times\lvert\cS^+\rvert}$ be a matrix such that $P_{(s,s')}=\cP(s'\lvert s)$ and $R \in \real^{\lvert\cS\rvert\times\lvert\cS\rvert}$ a diagonal matrix such that $R_{(s, s)} = e^{\eta\cR(s)}$. We also let ${\mathbf z}$ be the vector form of the value $z(s)$ for all states $s\in\cS$ and ${\mathbf z^+}$ an extended vector that also includes the known value $z(\tau)=e^{\eta\cJ(\tau)}$ for all terminal states $\tau\in\cT$. The problem is then expressed as
\begin{equation}
  {\bf z} = R P {\bf z^+}.
  \label{eq:eigen_lmdp}
\end{equation}
We can use the power iteration method over~\eqref{eq:eigen_lmdp} to obtain the solution for ${\bf z}$~\citep{Todorov2006}. Power iteration is guaranteed to converge as long as the diagonal matrix $R$ is not too large, and a common assumption is that the rewards of non-terminal states are non-positive (i.e.~$\cR(s)\leq 0$ for each $s\in\cS$). However, we refrain from making any such assumptions, and later we instead prove convergence in an alternative way.

Alternatively, when $\cP$ and $\cR$ are not known, the agent can learn an estimate $\hat z$ of the optimal value function in an online manner, using samples $(s_t, r_t, s_{t+1})$ generated when following the estimated policy $\hat\pi$ derived from $\hat z$ using \eqref{eq:optimal_policy}. The update rule for the so-called Z-learning algorithm is given by
\begin{align}
  \hat z(s_{t})&\gets(1-\alpha_t)  \hat z(s_{t}) + \alpha_t e^{\eta\cR(s_t, s_{t+1}, \hat\pi_t)} \hat z(s_{t+1}) \nonumber\\
  &\gets(1-\alpha_t)  \hat z(s_{t}) + \alpha_t e^{\eta r_t} \frac {\cP(s_{t+1}|s_t)} {\hat\pi_t(s_{t+1}|s_t)} \hat z(s_{t+1}).
  \label{eq:z_learning}
\end{align}
Here, $\alpha_t$ is a learning rate and $\cP(s_{t+1}|s_t)/\hat\pi_t(s_{t+1}|s_t)$ acts as an importance weight.


In the first-exit case, the solution of a set of \textit{component} problems can be combined to retrieve the optimal solution for new \textit{composite} problems with no further learning~\citep{Todorov2009a}. Assume we have a set of first-exit LMDPs ${\{\cL_i\}}_{i=1}^K$, which share $\cS, \cT$, $\cP$ and $\cR$, but differ in the values $z_i(\tau) = e^{\eta\cJ_i(\tau)}$ of terminal states. Let $z_1,\ldots,z_K$ be the optimal value functions of $\cL_1,\ldots,\cL_K$. Now consider a new composite problem $\cL$ that also shares the aforementioned elements with the component problems. If the value at terminal states can be expressed as a weighted sum as follows:
\[z(\tau) = \sum_{i=1}^K  w_i z_i(\tau) \;\;\forall\tau\in\cT, \] then by linearity of the value function, the same expression holds for non-terminal states~\cite{Todorov2009a}:
\[z(s) = \sum_{i=1}^K w_i z_i(s) \;\;\forall s\in\cS.\]

\subsection{Hierarchical Decomposition for LMDPs}

\citet{Infante2022} introduce a hierarchical decomposition for LMDPs. Given a first-exit LMDP ${\cL=\langle\cS,\cT,\cP,\cR,\cJ\rangle}$, the set of non-terminal states $\cS$ is partitioned into $L$ subsets $\{\cS_i\}_{i=1}^L$. Each subset $\cS_i$ induces a subtask $\cL_i=\langle\cS_i,\cT_i,\cP_i,\cR_i,\cJ_i\rangle$, i.e.~an LMDP for which
\begin{itemize}
  \item The set of non-terminal states is $\cS_i$.
  \item The set of terminal states $\cT_i=\{\tSi \in\cS^+\setminus\cS_i:\exists s\in \cS_i \; \text{s.t.} \; \cP(\tau|s)>0\}$ includes all states in $\cS^+\setminus\cS_i$ (terminal or non-terminal) that are reachable in one step from a state in $\cS_i$.
  \item $\cP_i:\cS_i\rightarrow\Delta(\cS_i^+)$ and $\cR_i:\cS_i\rightarrow\real$ are the restrictions of $\cP$ and $\cR$ to $\cS_i$, where $\cS_i^+=\cS_i\cup\cT_i$ denotes the full set of subtask states.
  \item The reward of a terminal state $\tSi \in\cT_i$ equals $\cJ_i(\tau)=\cJ(\tau)$ if $\tSi\in\cT$, and $\cJ_i(\tau)=\hat{v}(\tau)$ otherwise, where $\hat{v}(\tSi)$ is the estimated value in $\cL$ of the non-terminal state in  $\tSi\in\cS \setminus \cS_i$.
\end{itemize}

Intuitively, if the value $z_i(\tSi)$ of each terminal state $\tSi\in\cT_i$ equals its optimal value $z(\tSi)$ for the original LMDP~$\cL$, then solving the subtask $\cL_i$ yields the optimal values of the states in $\cS_i$.
In practice, however, the agent only has access to an estimate $\hat{z}(\tSi)$ of the optimal value.
In this case, the subtask $\cL_i$ is {\em parameterized} on the value estimate $\hat{v}$ (or $\hat{z}$) of the terminal states $\cT_i$, and each time the value estimate changes, the agent can solve $\cL_i$ to obtain a new value estimate
$\hat{z}(s)$ for each state $s\in\cS_i$.

The set of {\em exit states} $\cE=\cup_{i=1}^L\cT_i$ is the union of the terminal states of each subtask in $\{\cL_1,\ldots,\cL_L\}$. We let $\cE_i=\cE\cap\cS_i$ denote the set of (non-terminal) exit states in the subtask $\cL_i$. Also define the notation $K=\max_{i=1}^L|\cS_i|$, $N=\max_{i=1}^L|\cT_i|$ and $E=|\cE|$.

\begin{definition}
  Two subtasks $\cL_i$ and $\cL_j$ are equivalent if there exists a bijection $f:\cS_i\rightarrow\cS_j$ such that the transition probabilities and rewards of non-terminal states are equivalent through $f$.
\end{definition}
Using the above definition, we can define a set of equivalence classes $\cC=\{\cC_1,\ldots,\cC_C\}$, $C\leq L$, i.e.~a partition of the set of subtasks $\{\cL_1,\ldots,\cL_L\}$ such that all subtasks in a given partition are equivalent. Each equivalence class can be represented by a single subtask $\cL_j=\langle\cS_j,\cT_j,\cP_j,\cR_j,\cJ_j\rangle$. As before, the reward $\cJ_j$ of terminal states is parameterized on a given value estimate $\hat{v}$. We assume that each non-terminal state $s\in\cS$ can be mapped to its subtask $\cL_i$ and equivalence class $\cC_j$.

For each subtask $\cL_j$ and each terminal state $\tau^k\in\cT_j$, \citet{Infante2022}~introduce a {\em base LMDP} $\cL_j^k=\langle\cS_j,\cT_j,\cP_j,\cR_j,\cJ_j^k\rangle$ that shares all components with $\cL_j$ except the reward function on terminal states, which is defined as $\cJ_j^k(\tau)=1$ if $\tau=\tau^k$, and $\cJ_j^k(\tau)=0$ otherwise. Let $z_j^1,\ldots,z_j^n$ be the optimal value functions of the base LMDPs for $\cL_j$, with $n=|\cT_i|$. Given a value estimate $\hat z$ on each terminal state in $\cT_j$, due to compositionality we can express the value estimate of each state $s\in\cS_j$ as
\[
\hat z(s) = \sum_{k=1}^n \hat z(\tau^k) z_j^k(s).
\]
To solve the original LMDP, we can now define an optimal value function on exit states as $z_\cE:\cE\rightarrow\real$, and construct a matrix $G=\real^{|\cE|\times|\cE|}$ whose element $G_{(s,s')}$ equals $z_j^k(s)$ if $s'=\tau^k$ is the $k$-th terminal state in the subtask $\cL_j$ corresponding to the partition $\cS_i$ to which $s$ belongs, and $0$ otherwise. By writing ${\bf z}_\cE$ in vector form, the optimal value function satisfies the following matrix equation:
\[
{\bf z}_\cE = G {\bf z}_\cE.
\]
The total number of values required to represent the optimal value function equals $E + CKN$, since there are $C$ equivalence classes with at most $K$ states and $N$ base LMDPs.

\section{Average-reward Linearly-solvable Markov Decision Processes}

An average-reward Linearly-solvable Markov Decision Process (ALMDP) is a tuple
$\cL = \langle\cS,\cP,\cR\rangle$, where $\cS$ is a set of states, $\cP:\cS\rightarrow\Delta(\cS)$ is the passive dynamics, which describes state transitions in the absence of controls, and $\cR:\cS\rightarrow\real$ is a reward function. Since ALMDPs represent infinite-horizon tasks, there are no terminal states.

Throughout the paper, we make the following assumptions about ALMDPs.

\begin{assumption}
  The ALMDP $\cL$ is communicating~\citep{Puterman1994}: for each pair of states $s,s'\in\cS$, there exists a policy $\pi$ that has non-zero probability of reaching $s'$ from $s$.
  \label{ass:communicating}
\end{assumption}

\begin{assumption}
  The ALMDP $\cL$ is unichain~\citep{Puterman1994}: the transition probability distribution induced by all stationary policies admit a single recurrent class.
  \label{ass:unichain}
\end{assumption}

In the average-reward setting, the value function is defined as the expected average reward when following a policy $\pi:\cS\rightarrow\Delta(\cS)$ starting from a state $s\in\cS$. This is expressed as
\begin{equation}
  v_\eta^\pi(s) = \underset{T\rightarrow\infty}\lim \EEc{\frac{1}{T} \sum_{t=0}^T \cR(S_t, S_{t+1}, \pi)}{\pi, S_0 = s},
  \label{eq:value_function_almdp}
\end{equation}
where $\cR(s_t, s_{t+1}, \pi)$ is defined as for first-exit LMDPs.
Again, we are interested in obtaining the \textit{optimal} value function $v$. Under Assumption~\ref{ass:unichain}, the Bellman optimality equations can be written as
\begin{equation}
  v(s) = \frac 1 \eta \log{\sum_{s'\in\cS}\cP(s'\lvert s)\exp(\eta(\cR(s) - \rho + v(s')))} \;\;\forall s\in\cS,
  \label{eq:boe_almdp}
\end{equation}
where $\rho$ is the optimal one-step average reward (i.e.~gain), which is state-independent for unichain ALMDPs~\cite{Todorov2006}. Exponentiating yields
\begin{equation}
  z(s) = e^{\eta(\cR(s) - \rho)} \sum_{s'\in\cS}\cP(s'\lvert s)z(s') \;\;\forall s\in\cS.
  \label{eq:boe_z_almdp}
\end{equation}
For the optimal value function $z$, the optimal policy is given by the same expression as in~\eqref{eq:optimal_policy}.

\subsection{Solving an ALMDP}

We introduce the notation $\Gamma=e^{\eta\rho}$ (exponentiated gain). Similar to the first-exit case, Equation~\eqref{eq:boe_z_almdp} can be expressed in matrix form as
\begin{equation}\label{eq:rel_opt}
  \Gamma {\bf z} = R P {\bf z},
\end{equation}
where the matrices $P\in \real^{\lvert\cS\rvert\times\lvert\cS\rvert}$ and $R\in \real^{\lvert\cS\rvert\times\lvert\cS\rvert}$ are appropiately defined as in~\eqref{eq:eigen_lmdp}. The exponentiated gain $\Gamma$ can be shown to correspond to the largest eigenvalue of $RP$~\citep{Todorov2009}.
An ALMDP can be solved using {\em relative value iteration} by selecting a reference state $s^*\in\cS$, initializing $\hat{\bf z}_0={\bf 1}$ and iteratively applying
\begin{equation*}
  \hat{\bf z}_{k+\frac 1 2} \gets R P \hat{\bf z}_k, \quad \quad \hat{\bf z}_{k+1} \gets \hat{\bf z}_{k+\frac 1 2} / \hat z_{k+\frac 1 2}(s^*).
\end{equation*}
The reference state $s^*$ satisfies $z(s^*)=1$, which makes the optimal value $z$ unique (else any constant shift preserves optimality). After convergence, the exponentiated gain equals $\Gamma=\hat z_{k+\frac 1 2}(s^*)$. Under Assumption~\ref{ass:communicating}, relative value iteration converges to the unique optimal value $z$~\citep{Todorov2009}.

An alternative method for solving an ALMDP $\cL$ is to transform it to a first-exit LMDP. 
Given a reference state $s^*$, define a first-exit LMDP $\cL'=\langle\cS\setminus\{s^*\},\{s^*\},\cP',\cR',\cJ'\rangle$, where $\cP'(s'|s)=\cP(s'|s)$ for all state pairs $(s,s')\in(\cS\setminus\{s^*\})\times\cS$, and $\cJ(s^*)=0$ (implying $z(s^*)=1$). By inspection of~\eqref{eq:boe_z_lmdp} and~\eqref{eq:boe_z_almdp}, the Bellman optimality equation of $\cL'$ is identical to that of $\cL$ if $\cR'(s)=\cR(s)-\rho$. Even though the agent has no prior knowledge of the exponentiated gain $\Gamma = e^{\eta\rho}$, we can perform binary search to find $\Gamma$. For a given estimate $\hat\Gamma$ of $\Gamma$, after solving $\cL'$, we can compare $\hat \Gamma z(s^*)$ with $e^{\eta\cR(s^*)}\sum_s\cP(s|s^*)z(s)$. If $\hat \Gamma z(s^*)$ is greater, then $\hat\Gamma$ is too large, else it is too small.

Alternatively, when $\cP$ and $\cR$ are not known, we can obtain an estimate $\hat v$ of the optimal value $v$ and an estimate $\hat\rho$ of the optimal gain $\rho$ using \textit{differential soft TD-learning}, similar to differential Q-learning~\citep{Wan2021}. We collect samples $(s_t, r_t, s_{t+1})$ generated by the estimated policy $\hat\pi$ derived from $\hat v$ as in \eqref{eq:optimal_policy}, and derive update rules for $\hat v$ and $\hat \rho$ from \eqref{eq:boe_almdp} as follows:
\begin{align}
    \hat{v}_{t+1}(s_t) &\gets \hat{v}_t(s_t) + \alpha_t \delta_t,\label{eq:main_v_td_update}\\
    \hat{\rho}_{t+1} &\gets \hat{\rho}_t + \lambda \alpha_t \delta_t.\label{eq:main_rho_td_update}
    \end{align}
Here, the TD error $\delta_t$ is given by
\begin{align*}
\delta_t &= r_t - \hat{\rho}_t - \frac 1 \eta \log \frac {\hat{\pi}_t(s_{t+1}|s_t)} {\cP(s_{t+1}|s_t)} + \hat{v}_t(s_{t+1}) - \hat{v}_t(s_t)\\
 &= r_t - \hat{\rho}_t + \frac 1 \eta \log \sum_{s'\in\cS} \cP(s'|s_t) e^{\eta \hat{v}_t(s')} - \hat{v}_t(s_t).
\end{align*}
Note that both updates use the same TD error. At any time, we can retrieve estimates of $\hat z$ and $\widehat\Gamma$ by exponentiating $\hat v$ and $\hat \rho$, respectively.

\begin{theorem}
    Under mild assumptions, differential soft TD-learning in~\eqref{eq:main_v_td_update} and~\eqref{eq:main_rho_td_update} converges to the optimal values of $v$ and $\rho$ in $\cL$. \label{theo:almdps}
\end{theorem}

\begin{proof}[Proof sketch]
The proof is adapted from the proof of convergence of differential Q-learning~\cite{Abounadi2001,Wan2021}, which requires the ALMDP to be communicating (Assumption~\ref{ass:communicating}).
Define a Bellman operator $T$ as
\[
T(v)(s) = \cR(s) + \frac 1 \eta \log \sum_{s'\in\cS} \cP(s'|s) e^{ \eta v(s') }.
\]
To adapt the previous proof, it is sufficient to show that $T$ is a non-expansion in the max norm, i.e.~$\infnorm{T(x)-T(y)} \leq \infnorm{x - y}$ for each $x, y\in\real^{|\cS|}$, and that $T$ satisfies $T(x + c\mathbbm{1}) = T(x) + c\mathbbm{1}$ for each $x\in\real^{|\cS|}$ and constant $c\in\real$, where $\mathbbm{1}$ is the vector of $|\cS|$ ones.
For completeness, the full proof appears in Appendix~\ref{proof:theo_almdps}.
\end{proof}

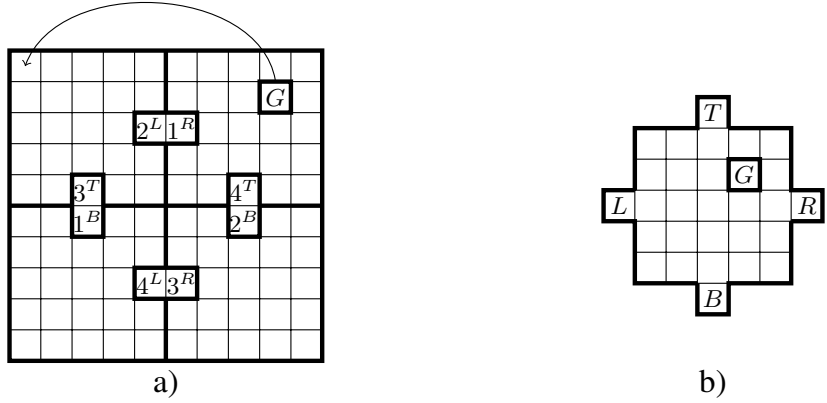
\begin{figure*}
  \begin{center}
    \begin{adjustbox}{width=0.6\textwidth}
      \begin{tikzpicture}
        \draw[step=0.4,thin,shift={(0.2,0.2)}] (0.8,0.8) grid (4.8,4.8);
        \draw[ultra thick] (1,1) rectangle (5,5);
        \draw[ultra thick] (3,1) -- (3,1.8);
        \draw[ultra thick] (3,2.2) -- (3,3.8);
        \draw[ultra thick] (3,4.2) -- (3,5);
        \draw[ultra thick] (1,3) -- (1.8,3);
        \draw[ultra thick] (2.2,3) -- (3.8,3);
        \draw[ultra thick] (4.2,3) -- (5,3);

        \draw[ultra thick] (4.2,4.2) rectangle (4.6,4.6);
        \draw[ultra thick] (3.8,2.6) rectangle (4.2,3.4);
        \draw[ultra thick] (1.8,2.6) rectangle (2.2,3.4);
        \draw[ultra thick] (2.6,3.8) rectangle (3.4,4.2);
        \draw[ultra thick] (2.6,1.8) rectangle (3.4,2.2);

        \node (R) at (1.2,4.8) {} ;
        \node (G) at (4.4,4.4) {\small $G$};
        \node at (2,3.2) {\small $3^T$};
        \node at (2,2.8) {\small $1^B$};
        \node at (4,3.2) {\small $4^T$};
        \node at (4,2.8) {\small $2^B$};
        \node at (2.8,4) {\small $2^L$};
        \node at (2.8,2) {\small $4^L$};
        \node at (3.2,4) {\small $1^R$};
        \node at (3.2,2) {\small $3^R$};

        \draw[step=0.4,thin,shift={(0.2,0)}] (8.799,1.999) grid (10.8,4);
        \draw[ultra thick] (9,3.2) -- (8.6,3.2) -- (8.6,2.8) -- (9,2.8) -- (9,2) -- (9.8,2);
        \draw[ultra thick] (9,3.2) -- (9,4) -- (9.8,4) -- (9.8,4.4) -- (10.2,4.4) -- (10.2,4);
        \draw[ultra thick] (10.2,4) -- (11,4) -- (11,3.2) -- (11.4,3.2) -- (11.4,2.8) -- (11,2.8);
        \draw[ultra thick] (9.8,2) -- (9.8,1.6) -- (10.2,1.6) -- (10.2,2) -- (11,2) -- (11,2.8);
        \draw[ultra thick] (10.2,3.2) rectangle (10.6,3.6);

        \node at (10.4,3.4) {\small $G$};
        \node at (8.8,3)    {\small $L$};
        \node at (11.2,3)   {\small $R$};
        \node at (10,1.8)   {\small $B$};
        \node at (10,4.2)   {\small $T$};

        \node at (3,0.7) {\Large a)};
        \node at (10,0.7) {\Large b)};

        \draw [->] (G.north) to [out=100,in=70] (R.center);
      \end{tikzpicture}
    \end{adjustbox}
  \end{center}
  \caption{a) An example $4$-room ALMDP; b) a single subtask with 5 terminal states $G,L,R,T,B$ that is equivalent to all 4 room subtasks. Rooms are numbered 1 through 4, left-to-right, then top-to-bottom, and exit state $1^B$ refers to the exit $B$ of room $1$, etc. \\}
  \label{fig:ex}
\end{figure*}

\section{Hierarchical Average-Reward LMDPs}

In this section we present our approach for hierarchical average-reward LMDPs. The idea is to take advantage of the similarity of the value functions in the first-exit and average-reward settings, and use compositionality to compose the value functions of the subtask LMDPs without additional learning.

\subsection{Hierarchical Decomposition}
Consider an ALMDP $\langle\cS,\cP,\cR\rangle$. 
Similarly to~\citet{Infante2022}, we assume that the state space $\cS$ is partitioned into subsets $\left\{\cS_i\right\}^L_{i=1}$, with each partition $\cS_i$ inducing
a first-exit LMDP $\cL_i = \langle\cS_i, \cT_i, \cP_i, \cR_i,\cJ_i\rangle$.
The components of each such subtask $\cL_i$ are defined as follows:
\begin{itemize}
  \item The set of states is $\cS_i$.
  \item The set of terminal states $\cT_i=\{\tau\in\cS\setminus\cS_i: \exists s\in\cS_i, \cP(\tau|s)>0\}$ contains states not in $\cS_i$ that are reachable in one step from any state inside the partition.
  \item The transition function $\cP_i$ and reward function $\cR_i$ are projections of $\cP$ and $\cR-\hat\rho$ onto $\cS_i$, where $\hat\rho$ is a gain estimate.
  \item $\cJ_i$ is defined for each $\tau\in\cT_i$ as $\cJ_i(\tau)=\hat v(\tau)$, where $\hat v$ is a current value estimate (hence $z_i(\tau)=e^{\eta\hat v(\tau)} = \hat z(\tau)$ is defined by a current exponentiated value estimate $\hat z$).
\end{itemize}
The Bellman optimality equations of each subtask $\cL_i$ are given by
\begin{equation}\label{eq:subtask}
  z_i(s) = e^{\eta \cR_i(s)} \sum_{s'} \cP_i(s'|s) z_i(s') \;\; \forall s\in\cS_i.
\end{equation}
By inspection of the Bellman optimality equations in~\eqref{eq:boe_z_almdp} and~\eqref{eq:subtask}, they are equal if $\cR_i(s)=\cR(s)-\rho$. Thus, if $z_i(\tau)=z(\tau)$ for each $\tau\in\cT_i$ then the solution of the subtask $\cL_i$ corresponds to the optimal solution for each $s\in\cS_i$. However, in general neither $\rho$ nor $z(\tau)$ are known prior to learning and, therefore, we have to use estimates $\hat\rho$ and $\hat z(\tau)$. Each subtask $\cL_i$ can be seen as being {\it parameterized\/} on the value estimates $\hat z(\tau)$ for each $\tau\in\cT_j$ and the gain estimate $\hat\rho$. Every time that $\hat z(\tau)$, $\tau\in\cT_i$, and $\hat\rho$ change, we obtain a new value estimate for each $s\in\cS_i$ by solving the subtask for the new parameters.

\subsection{Subtask Compositionality}
It is impractical to solve each subtask $\cL_i$ every time the estimate $\hat z(\tau)$ changes for $\tau\in\cT_j$. To alleviate this computation we leverage compositionality for LMDPs. The key insight is to build a basis of value functions that can be combined to obtain the solution for the subtasks.

Consider a subtask $\cL_i=\langle\cS_i,\cT_i,\cP_i,\cR_i,\cJ_i\rangle$ and let $n=|\cT_i|$.  We introduce $n$ base LMDPs $\{\cL_i^1,\ldots,\cL_i^n\}$ that are first-exit LMDPs and terminate in $\cT_i$. These base LMDPs only differ from $\cL_i$ in the reward of each terminal state $\tau^k\in\cT_i$. For all $s\in\cS_i$, the reward for each $\cL_i^k$ is by definition $\cR_i(s)=\cR(s)-\hat\rho$ for all $s\in\cS_i$, while at terminal states $\tau\in\cT_i$ we let the reward function be $z_i^k(\tau;\hat\rho)=1$ if $\tau=\tau^k$ and $z_i^k(\tau;\hat\rho)=0$ otherwise. Thus, the base LMDPs are parameterized by the gain estimate $\hat\rho$. This is equivalent to setting the reward to $\cJ_i^k(\tau)=0$ if $\tau=\tau_k$ and $\cJ_i^k(\tau)=-\infty$ otherwise. Intuitively, each base LMDP solves the subtask of reaching one specific terminal state $\tau_k\in\cT_i$.

Let us now assume that we have the solution $z_i^1(\cdot;\rho),\ldots,z_i^n(\cdot;\rho)$ for the base-LMDPs (for the optimal gain $\rho$) as well as the optimal value $z(\tau^k)$ of the original ALMDP for each terminal state $\tau^k\in\cT_i$. Then by compositionality we could represent the value function of each terminal state as a weighted combination of the subtasks:
\begin{equation}
  z(\tau) = \sum_{k=1}^n w_k z_i^k(\tau;\rho) =  \sum_{k=1}^n z(\tau^k) z_i^k(\tau;\rho) \;\;\forall\tau\in\cT_i.
  \label{eq:comp_terminal}
\end{equation}

Clearly, the RHS in the previous expression evaluates to $z(\tau)$ since $z(\tau^k) z_i^k(\tau;\rho) = z(\tau)\cdot 1$ when $\tau = \tau^k$, and
$z(\tau^k) z_i^k(\tau;\rho) = z(\tau^k)\cdot 0$ otherwise.

Thanks to compositionality, we can also represent the value function for each subtask state $s\in\cS_i$ as
\begin{equation}
  z(s) = \sum_{k=1}^n z(\tau^k) z_i^k(s;\rho)\;\;\forall s\in\cS_i.
  \label{eq:comp_internal}
\end{equation}
We remark that the base LMDPs depend on the gain $\rho$ by the definition of the reward function. This parameter is not known prior to learning. The subtasks in practice are solved for the latest estimate $\hat\rho$ and must be re-learned for every update of this parameter until convergence.
\subsection{Efficiency of the value representation}
 Similar to previous work~\citep{Wen2020,Infante2022} we can exploit the equivalence of subtasks to learn more efficiently. Let $\cC=\{\cC_1,\ldots,\cC_C\}$, $C\leq L$, be a set of equivalence classes, i.e.~a partition of the set of subtasks $\{\cL_1,\ldots,\cL_L\}$ such that all subtasks in a given partition are equivalent.
As before, we also define a set of exit states as $\cE=\cup_{i=1}^L\cT_i$.
Due to the decomposition, we do not need to keep an explicit value estimate $\hat z(s)$ for every state $s\in\cS$. Instead, it is sufficient to keep a value function for exit states $\hat z_\cE: \cE\rightarrow\real$ and a value function for each base LMDP of each equivalence class. This is enough to represent the value for any state $s\in\cS$ using the compositionality expression in~\eqref{eq:comp_internal}.

Letting $K=\max_{i=1} ^L\lvert\cS_i\rvert$,  $N=\max_{i=1} ^L\lvert\cT_i\rvert$ and $E=\lvert\cE\rvert$, $O(KN)$ values are needed to represent the base LMDPs of a subtask, and we can thus represent the value function with $O(CKN + E)$ values. The decomposition leads to an efficient representation of the value function whenever $CKN + E \ll \lvert\cS\rvert$. This is achieved when there are few equivalence classes, the size of each subtask is small (in terms of the number of states) and there are relatively few exit states.


  \emph{Example 1}:
  Figure~\ref{fig:ex}a) shows an example 4-room ALMDP. When reaching the state marked $G$, separate from the room but reachable in one step from the highlighted location, the system transitions to a restart state (top left corner) and receives reward $0$. In all other states the reward is $-1$. The rooms are connected via doorways, so the subtask corresponding to each room has two terminal states in other rooms, plus the terminal state $G$ in the top right room. The 9 exit states in $\cE$ are highlighted and correspond to states next to doorways, plus $G$. Figure~\ref{fig:ex}b) shows a single subtask that is equivalent to all four room subtasks, since the dynamics is shared inside rooms and the set of terminal states is the same. 
    There are five base LMDPs with value functions $z^G$, $z^L$, $z^R$, $z^T$ and $z^B$, respectively. Given an initial value estimate $\hat{z}_\cE$ for each exit state in $\cE$, a value estimate of any state in the top left room is given by $\hat{z}(s)=\hat{z}_\cE(1^B) z^B(s) + \hat{z}_\cE(1^R) z^R(s)$, where we use $\hat{z}_\cE(G)=\hat{z}_\cE(L)=\hat{z}_\cE(T)=0$ to indicate that the terminal states $G$, $L$ and $T$ are not reachable in the top left room. 
    The total number of values needed to store the optimal value function is $E+CKN=9+125=134$, and the base LMDPs are faster to learn since they have smaller state space.

\begin{figure*}[!t]
  \begin{center}
  \includegraphics*[width=0.32\textwidth]{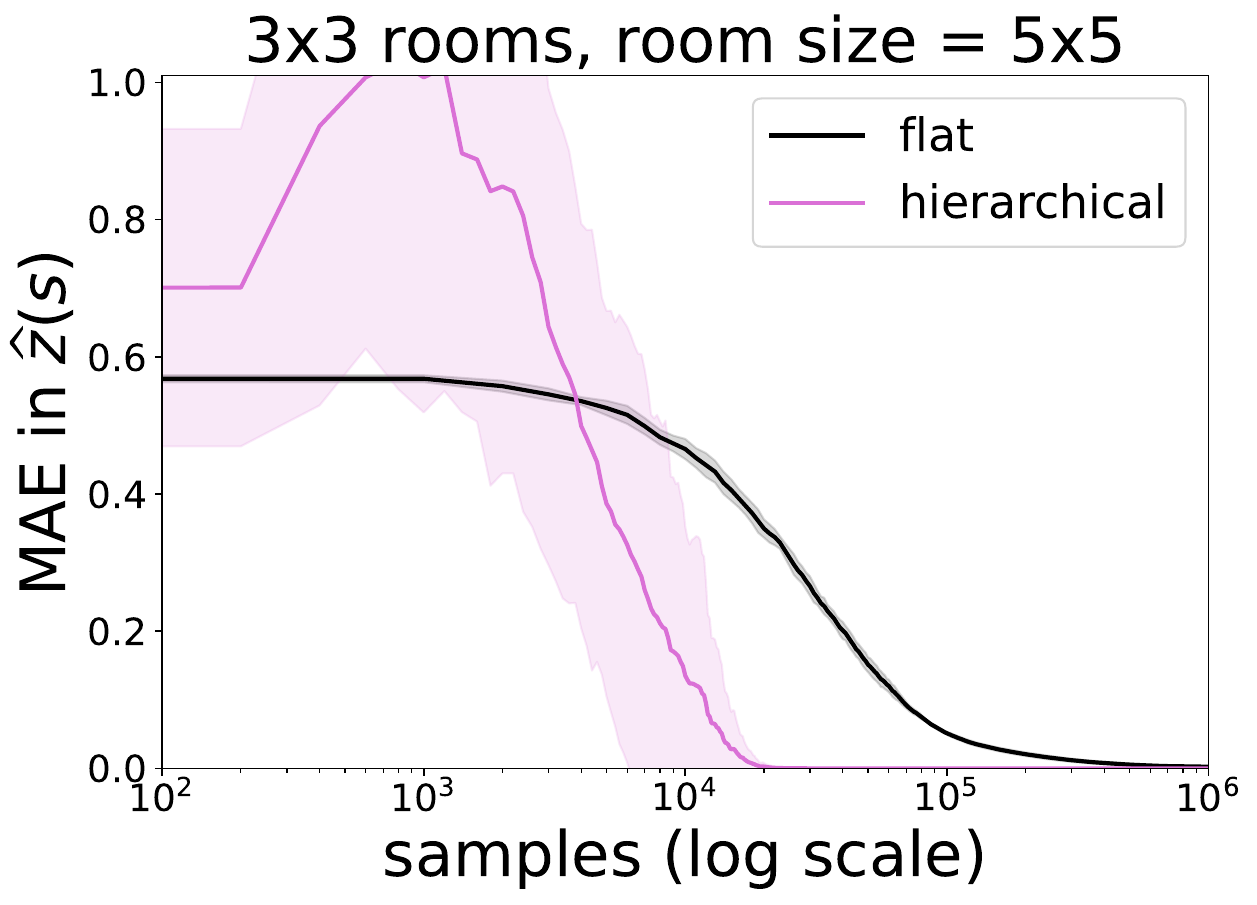}
  \includegraphics*[width=0.32\textwidth]{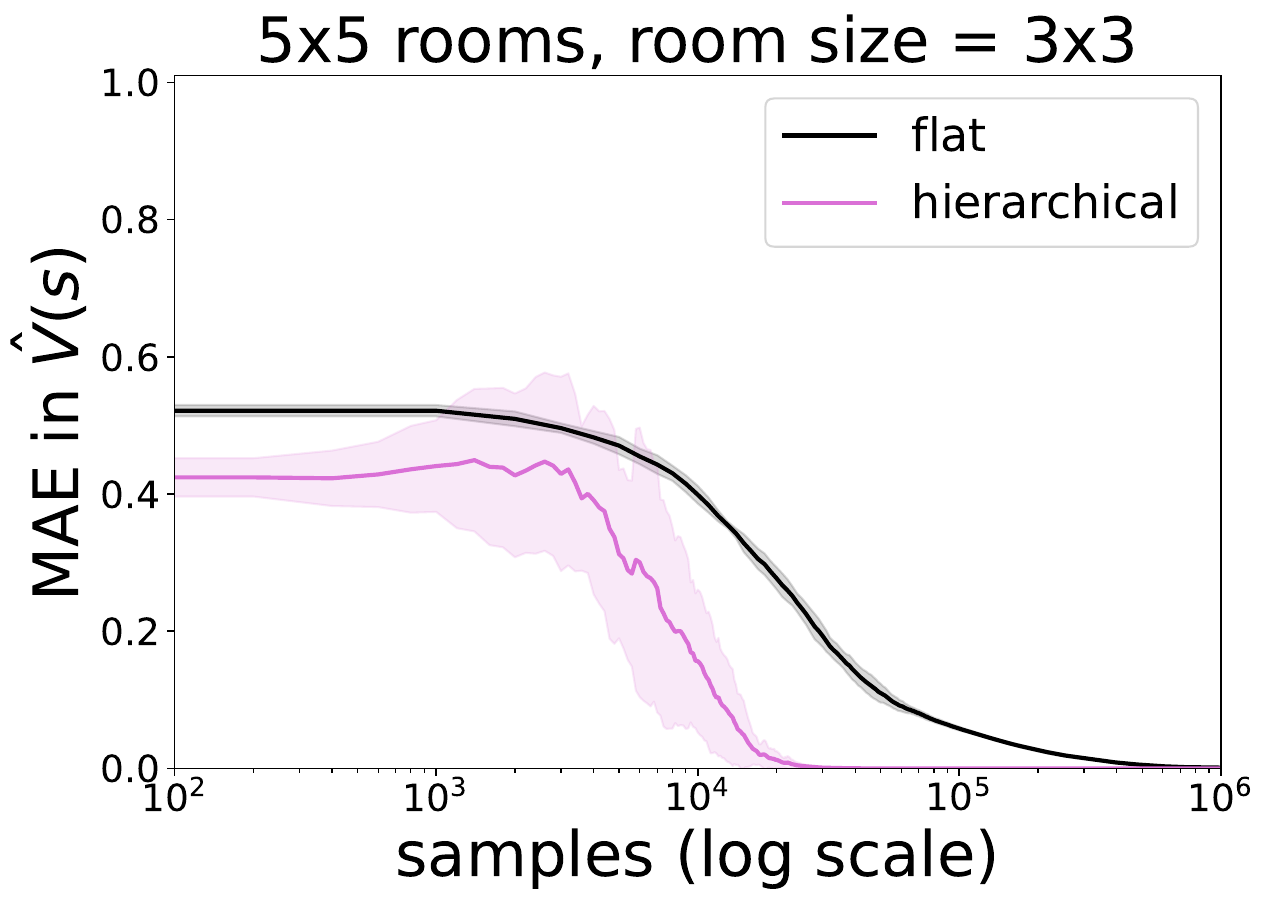}
  \includegraphics*[width=0.32\textwidth]{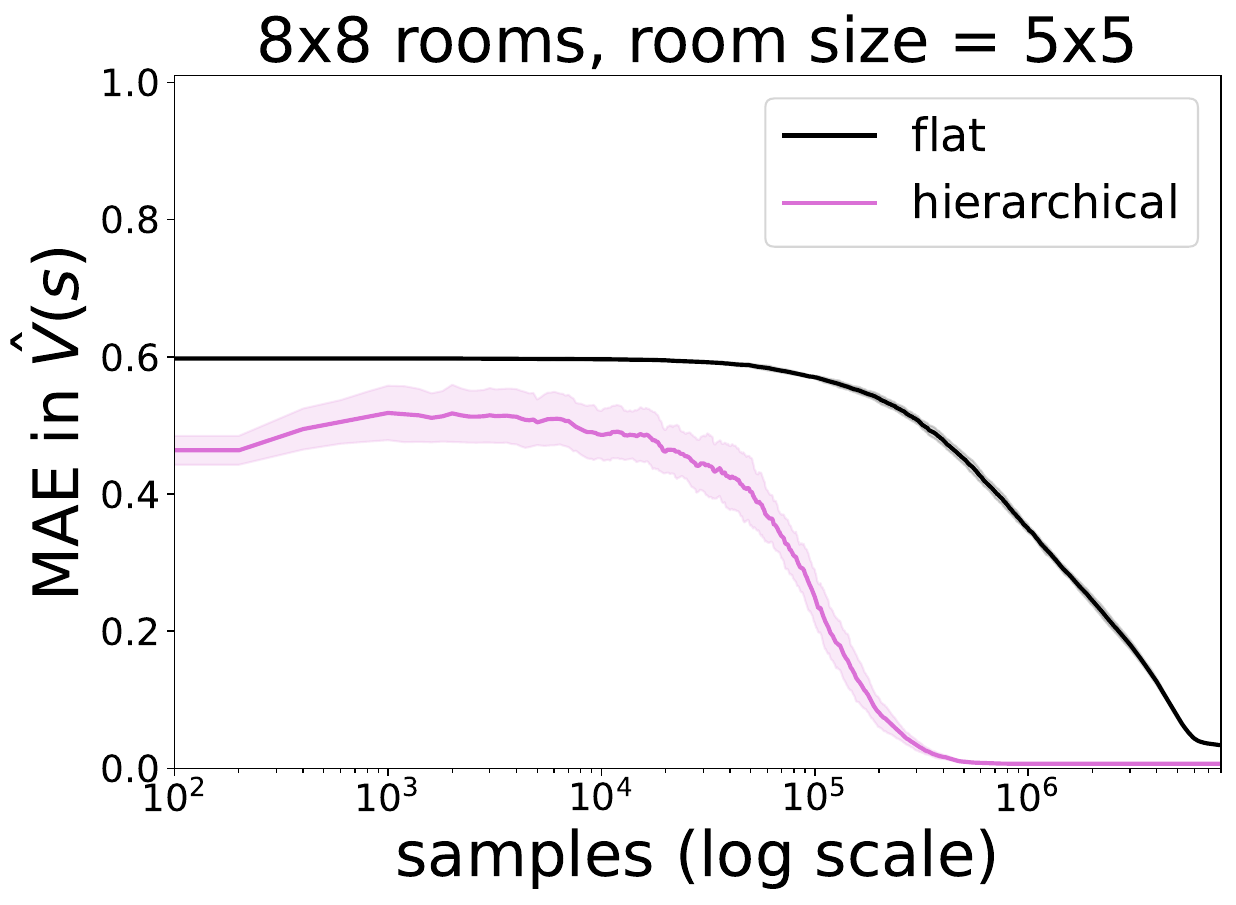}
  \caption{Results in N-room when varying the number of rooms and the size of the rooms.}
  \label{fig:nrooms}
  \end{center}
\end{figure*}

\section{Algorithms}
We now propose two algorithms for solving hierarchical ALMDPs. The first is a two-stage eigenvector approach that relies on first solving the subtasks. The second is an online algorithm in which an agent simultaneously learns the subtasks, the gain and the exit values from samples $(s_t, r_t, s_{t+1})$.
Once again we recall that we do not explicitly represent the values for states $s\notin\cE$.

\subsection{Eigenvector approach}

In previous work, the base LMDPs are only solved once, and the solutions are then reused to compute the value function $z_\cE$ on exit states. However, in the case of ALMDPs, the reward functions of base LMDPs depend on the current gain estimate $\hat\rho$, which is initially unknown. 



Our proposed algorithm appears in Algorithm~\ref{alg:eigenvector}. The intuition is that in each iteration, we first solve the subtasks for the latest estimate of the exponentiated gain $\widehat\Gamma$. For this, we use~\eqref{eq:subtask} with the current value of $\hat\rho$ to solve the base LMDPs. We then apply~\eqref{eq:comp_internal} restricted to $\cE$ to obtain an estimate of the value for the exit states. This yields the system of linear equations
\begin{equation}
    {\bf z_\cE} = G_\cE {\bf z_\cE}.\label{eq:eigenvector}
\end{equation}
Here, the matrix $G_\cE\in\real^{\lvert\cE\rvert\times \lvert\cE\rvert}$ contains the optimal values of the base LMDPs and has elements defined as in~\eqref{eq:comp_internal}. We use the previously introduced idea to transform the ALMDP $\cL$ to a first-exit LMDP $\cL'$ parameterized on the estimated gain $\hat\rho$, and find the optimal exponentiated gain $\Gamma$ using binary search. We keep a reference state $s^*\in\cS$ (which is by definition an exit state) and use the test described above to decide how to update the search interval. 
\begin{algorithm}[!b]
  \caption{Eigenvector approach to solving a hierarchical ALMDP.}
  \begin{algorithmic}[1]

    \State{{\bf Input:} $\cL,\cS_1,\ldots,\cS_L,\cE,\epsilon,\eta$}
    \State $\text{lo}\gets 0$, $\text{hi}\gets 1$
    \While {$\text{hi} - \text{lo} > \epsilon$}
    \State $\widehat\Gamma \gets (\text{hi} + \text{lo}) \mathbin{/} 2$
    \State Solve base LMDPs $\cL_j^1,\ldots,\cL_j^n$ for each equivalence class $\cC_j$
    \State Form the matrix $G_\cE$ from the optimal value functions
    \State Solve the system of equations  ${\bf \hat z_\cE} = G_\cE {\bf\hat z_\cE}$
    \If {$ \widehat\Gamma \hat z_\cE(s^*) > e^{\eta \cR(s^*)} \sum_{s\in\cS} \cP(s\lvert s^*) \hat z_\cE(s)$}
    \State $\text{hi}\gets \widehat\Gamma$
    \Else \State $\text{lo}\gets \widehat\Gamma$
    \EndIf
    \vspace*{3pt}
    \EndWhile
    \Return value functions of all base LMDPs, ${\bf \hat z_\cE}$
  \end{algorithmic}
  \label{alg:eigenvector}
\end{algorithm}

\begin{theorem}\label{thm:converge}
    Algorithm~\ref{alg:eigenvector} converges to the optimal value function $z$ of $\cL$ as $\epsilon\to 0$.
\end{theorem}

First note that the optimal value function $z$ of $\cL$ exists and is unique due to Assumption~\ref{ass:communicating}. Due to the equivalence between $\cL$ and the corresponding first-exit LMDP $\cL'$, this implies that $\cL'$ has a unique solution $z(\cdot;\rho)$ when the estimated gain $\hat\rho$ equals $\rho$, and that this solution equals $z(\cdot;\rho)=z$, the optimal solution to $\cL$.

\begin{lemma}\label{lemma:monotonicity}
     Given a first-exit LMDP $\cL'$ parameterized on $\hat\rho$, the optimal value $z(s;\hat\rho)$ of each non-terminal state $s\in\cS$ is strictly monotonically decreasing in $\hat\rho$.
\end{lemma}

\begin{proof}
Strict monotonicity requires that there exists $\varepsilon>0$ such that ${z(s;\hat\rho - \varepsilon) > z(s;\hat\rho) > z(s;\hat\rho+\varepsilon)}$ when $\varepsilon\rightarrow 0$. We prove the first inequality by induction; the second is analogous. The base case is given by the terminal states $\tau\in\cT$, for which $z(\tau;\hat\rho - \varepsilon) = z(\tau;\hat\rho)$. The inductive case is given by
\begin{align*}
    z(s;\hat\rho-\varepsilon) &= e^{\eta(\cR(s) - (\hat\rho -\varepsilon))}\sum_{s'\in\cS}\cP(s'\lvert s) z(s';\hat\rho- \varepsilon)\\
      &\geq e^{\eta\varepsilon} e^{\eta(\cR(s) - \rho))}\sum_{s'\in\cS}\cP(s'\lvert s) z(s';\hat\rho)\\
      &= e^{\eta\varepsilon} z(s;\hat\rho) > z(s;\hat\rho).
\end{align*}
This concludes the proof.
\end{proof}

As a consequence of Lemma~\ref{lemma:monotonicity}, $\cL'$ has a unique solution $z(\cdot,\hat\rho)$ for each $\hat\rho\geq\rho$, since the values $z(\cdot,\hat\rho)$ decrease as $\hat\rho$ increases. In contrast, there may be values of $\hat\rho>\rho$ for which power iteration does not converge.

\begin{restatable}{lemma}{optimality}
Given a subtask $\cL_i$, if the optimal value of each terminal state $\tau\in\cT_i$ equals its optimal value in $\cL$, i.e. $z_i(\tau) = z(\tau)$, and the optimal gain $\rho$ in $\cL$ is known, then the optimal value of each non-terminal state $s\in\cS_i$ is unique and equals $z_i(s)=z(s)$.
    \label{lemma:optimality}
\end{restatable}

\begin{proof} Since $\cR_i$ and $\cP_i$ are restrictions of $\cR-\rho$ and $\cP$, respectively, to $\cS_i$, we have
\begin{align*}
    z_i(s) &= e^{\eta\cR_i(s)}\sum_{s'}\cP_i(s'\lvert s) z_i(s') = e^{\eta(\cR(s) - \rho)}\sum_{s'}\cP(s'\lvert s) z_i(s'), \nonumber
\end{align*}
which is the same Bellman equation as for $z(s)$. Assuming that $z_i(\tau) = z(\tau)$ for each $\tau\in\cT$, directly yields that $z_i(s)=z(s)$ for each non-terminal state $s\in\cS_i$.
\end{proof}

\begin{corollary}\label{cor:uniqueness}
    If the optimal gain $\rho$ in $\cL$ is known, each base LMDP $\cL_j^k$ has a unique solution $z_j^k(\cdot;\rho)$.
\end{corollary}

\begin{proof} From \eqref{eq:comp_internal}, the optimal values of subtask states satisfy
\[
  z(s) = \sum_{k=1}^n z(\tau^k) z_i^k(s;\rho)\;\;\forall s\in\cS_i.
\]
Due to Lemma~\ref{lemma:optimality}, the optimal value $z(s)$ is unique, which is only possible if $z_i^k(s;\rho)$ is unique for each $\tau^k\in\cT_i$.
\end{proof}

Combined with Lemma~\ref{lemma:monotonicity}, Corollary \ref{cor:uniqueness} implies that each base LMDP $\cL_j^k$ has a unique solution $z_j^k(\cdot;\hat\rho)$ whenever $\hat\rho\geq\rho$.

\begin{lemma}
For $\hat\rho\geq\rho$, the equation ${\bf \hat z_\cE} = G_\cE {\bf\hat z_\cE}$ has a unique solution that equals $z_\cE(\tau)=z(\tau;\hat\rho)$ for each exit $\tau\in\cE$, where $z(\cdot;\hat\rho)$ is the unique value of the first-exit LMDP $\cL'$ for $\hat\rho$.
\end{lemma}

\begin{proof}
At convergence, due to \eqref{eq:comp_internal} it has to hold for each non-terminal exit $\tau\in\cE$ that
\[
  z_\cE(\tau) = \sum_{k=1}^n z_\cE(\tau^k) z_i^k(s;\hat\rho)\;\;\forall s\in\cS_i,
\]
where each $\tau^k$ is also an exit and $z_i^k(s;\hat\rho)$ is well-defined and unique since $\hat\rho\geq\rho$. This equation is satisfied when $z_\cE(\tau)=z(\tau;\hat\rho)$ for each exit. Since $z(\cdot;\hat\rho)$ is unique, this is the only solution.
\end{proof}

We now have all the ingredients to prove Theorem~\ref{thm:converge}. When $\hat\rho\geq\rho$ (or equivalently, $\widehat\Gamma\geq\Gamma$), each base LMDP has a unique solution, and $z_\cE$ is also unique. Moreover, when $\hat\rho>\rho$, the condition on line 8 is true, which causes binary search to discard all values greater than $\hat\rho$. If the base LMDPs or $z_\cE$ do not have a unique solution, we know that $\hat\rho$ is too small, and we can hence discard all values less than $\hat\rho$. Since the solution $z(\cdot;\hat\rho)$ is monotonically decreasing in $\hat\rho$, binary search is guaranteed to find the optimal gain $\rho$ within a factor of $\epsilon$.

\subsection{Online algorithm}
 In the online case (see Algorithm~\ref{alg:online}), the agent keeps an estimate of the exponentiated gain $\widehat\Gamma=e^{\eta\hat\rho}$ which is updated every timestep. It also keeps estimates of the value functions of the base LMDPs $\hat z_i^1(\cdot;\hat\rho),\ldots,\hat z_i^n(\cdot;\hat\rho)$ for each equivalence class $\cC_i$, and estimates of the value function on exit states $\hat z_\cE$.  All the base LMDPs of the same equivalence class can be updated with the same sample using intra-task learning with the appropriate {\it importance sampling weights\/}~\citep{Jonsson2016}. For the estimates of the exit states, we only update them upon visitation of such states. In that case, we use the compositionality expression in~\eqref{eq:comp_internal} to derive the following update:
\begin{equation}
  \hat z_\cE(s)\gets (1-\alpha_\ell) \hat z_\cE(s) + (1-\alpha_\ell) \sum_{k=1}^n \hat z_\cE(\tau^k) \hat z_i^k(s;\rho).
  \label{eq:td_update_exit}
\end{equation}
Here, $\alpha_\ell$ is the learning rate. Each of the learned components (i.e., gain, base LMDPs and exit state value estimates) maintain independent learning rates.

\begin{algorithm}[!htpb]
  \caption{Online algorithm.}
  \begin{algorithmic}[1]
    \State{{\bf Input:} $\cL,\cS_1,\ldots,\cS_L,\cE,s_0,\epsilon,\eta$}
    \State $t \gets 0$, $\widehat{\Gamma} \gets 1$
    \While {{\it not terminate}}
    \State Observe $(s_{t}, r_{t}, s_{t+1})\sim\hat\pi_t$
    \State Update $\hat z_j^1(\cdot;\hat\rho),\ldots,\hat z_j^n(\cdot;\hat\rho)$ using Equation~\eqref{eq:main_v_td_update}
    \State Compute $\hat z(s_t)$ and $\sum_{s'}\cP(s'|s_t)\hat z(s')$ using Equation \eqref{eq:comp_internal}
    \State Update $\widehat\Gamma$ using Equation~\eqref{eq:main_rho_td_update}
    \If{$s_t\in\cE$}
    \State Update $\hat z_\cE(s_t)$ using Equation~\eqref{eq:td_update_exit}
    \EndIf
    \EndWhile
  \end{algorithmic}
  \label{alg:online}
\end{algorithm}

\section{Experiments}

In this section we compare experimentally\footnote{Code available at \url{https://github.com/guillermoim/halmdps}} Algorithm~\ref{alg:online} with differential soft TD-learning in the flat representation of the ALMDP. We measure the Mean Absolute Error (MAE) between the estimated value function $\hat z$ and the true optimal value function $z$. For each algorithm, we report 
mean and standard deviation on five seeds. The learning rates have been optimized independently for each of the instances. We adapt two episodic benchmark tasks~\citep{Infante2022} and transform them into infinite-horizon tasks as follows:

  \noindent{\bf N-room domain}, cf.~Example 1. As in the episodic case, there are some `goal' states with high reward (i.e.~0). When the agent enters a goal state, the next action causes it to receive the given reward and transition to a {\it restart\/} location. We vary the size of the rooms as well as the size of the rooms.

  \noindent{\bf Taxi domain}. In this variant of the original domain~\citep{Dietterich2000}, once the passenger has been dropped off, the system transitions to a state in which the driver is in the last drop-off location, but a new passenger appears randomly at another location. 
  
\begin{figure}[!hb]
  \begin{center}
      \includegraphics*[width=0.32\textwidth]{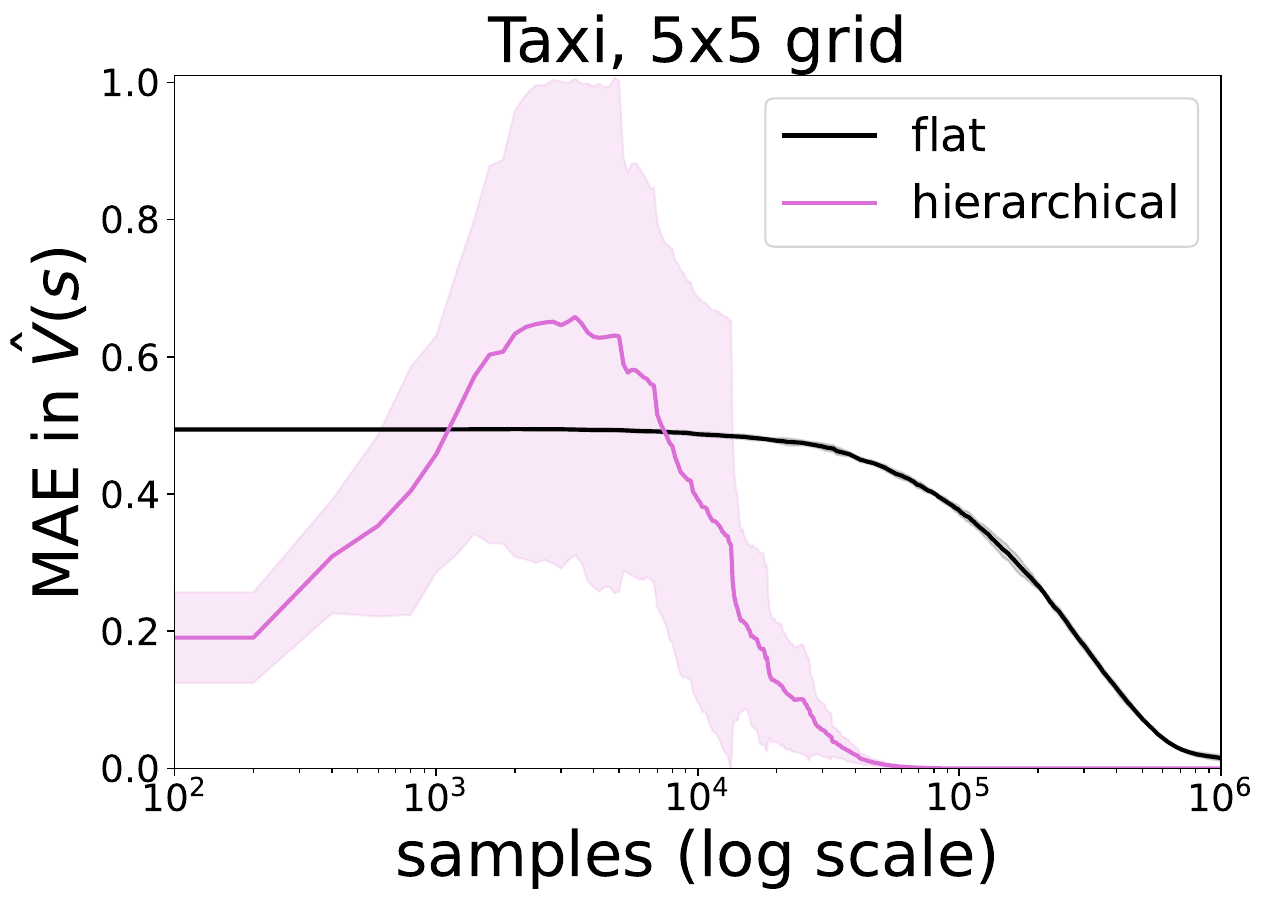}
      \includegraphics*[width=0.32\textwidth]{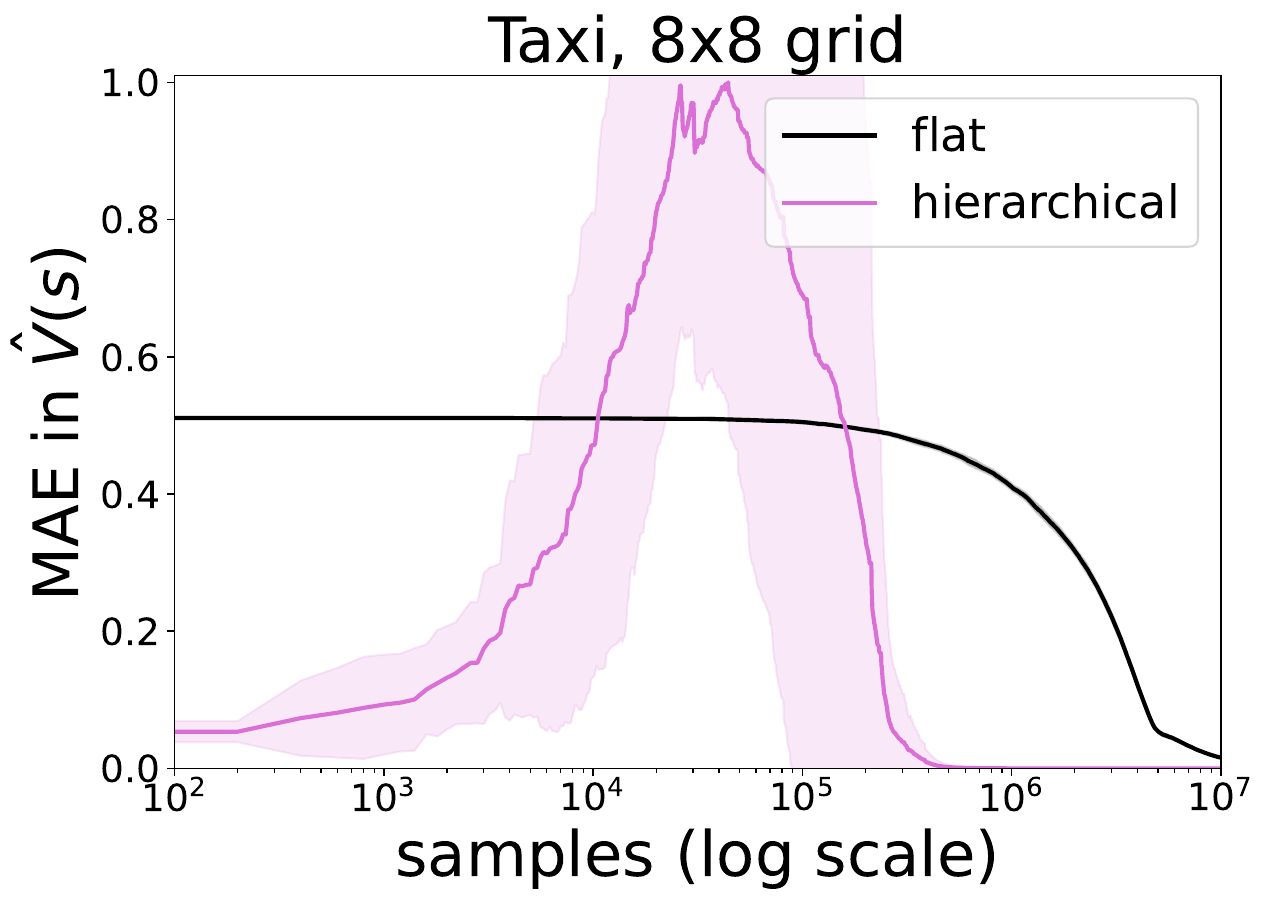}
      \caption{Results for $5 \times 5$ (top) and $8 \times 8$ (bottom) grids of the Taxi domain.}
        \label{fig:taxi}
  \end{center}
\end{figure}

Figures~\ref{fig:nrooms} and~\ref{fig:taxi} show the results. Our algorithm is able to speed up 
learning and converges to the optimal solution faster than flat average-reward reinforcement learning (note the log scale). This is in line with previous results for the episodic case~\citep{Infante2022}. The difference in the error scale in the figures is due to the initialization of the base LMDPs. The average reward setting poses an extra challenge since the `sparsity' of the reward can make the estimates of the gain oscillate. This ultimately has an impact on the estimates of the base LMDPs and the value estimates of the exit states, and it is likely the reason why in Figure~\ref{fig:taxi} the error increases before decreasing down to zero.



\section{Conclusion}

In this paper we present a novel framework for hierarchical average-reward reinforcement learning which makes it possible to learn the low-level and high-level tasks simultaneously. 
We propose an eigenvector approach and an online algorithm for solving problems in our framework, and show that the former converges to the optimal value function. As a by-product of our analysis, we also provide a convergence theorem in the non-hierarchical case for average-reward LMDPs, which to the extent of our knowledge, was not previously done. In the future we would like to prove convergence also for the proposed online algorithm.

\begin{ack}
Anders Jonsson is partially funded by TAILOR (EU H2020 \#952215), AGAUR SGR and Spanish grant PID2019-108141GB-I00. This publication is part of the action CNS2022-136178 financed by MCIN/AEI/10.13039/501100011033 and by the European Union Next Generation EU/PRTR.
\end{ack}


\bibliography{mybibfile}

\newpage
\appendix
\onecolumn

\section{Proof Theorem~\ref{theo:almdps}}\label{proof:theo_almdps}
\subsection{Preliminaries}
We introduce the notation:
\begin{itemize}
    \item $\allones$ denotes an all-ones vector of length $\lvert\cS\rvert$.
    \item $\indicator{p}$ is the indicator function that takes $1$ when predicate $p$ is true and $0$ otherwise.
\end{itemize}

We assumme an underlying continuing LMDP $\cL=\langle\cS,\cP,\cR\rangle$ where $\cS$ represents the state space, $\cP$ the passive dynamics and $\cR$ the reward function. Similarly to [Section B.1] in~\cite{Wan2021}, we also assume there exists a set-valued process $\{X_t\}$ where $X_t$ is a non-empty subset defined as ${X_t = \{ (s) : s\;\text{component of $v$ was updated at timestep $t$} \}}$.

We recall that the TD updates in the asynchronous case are 
\begin{align}
\widehat{v}_{t+1}(s) &\gets \widehat{v}_t(s) + \alpha_t(s) \delta_t(s) \indicator{s\in X_t},\label{eq:updatev}\\
\widehat{\rho}_{t+1} &\gets \widehat{\rho}_t + \lambda \sum_s \alpha_t(s) \delta_t(s) \indicator{s\in X_t}.\label{eq:updaterho}
\end{align}
The indicator $\indicator{s\in X_t}$ specifies whether the value of state $s$ updates at timestep $t$. The TD error for state $s$ is
\begin{align*}
\delta_t(s) 
 &= r_t(s) - \widehat{\rho}_t + \frac 1 \eta \log \sum_{s'\in\cS} \cP(s'|s) e^{\eta \widehat{v}_t(s')} - \widehat{v}_t(s).
\end{align*}

We introduce a series of necessary assumptions for convergence. We adapt Assumptions B.1-B.5 in~\cite{Wan2021} to the case of LMDPs. Assumptions~\ref{ass:communicating} and~\ref{ass:uniqueness} are standard in average-reward settings, while Assumption~\ref{ass:stepsize1} is the standard Robbins-Monro conditions for step sizes. Assumptions~\ref{ass:stepsize2} and~\ref{ass:stepsize3} are introduced in the convergence argument of RVI Q-learning by~\citet{Borkar1998} and specify some requirements for the learning rates when asynchronous updates are performed. For more details we refer the reader to Section B.1 of~\cite{Wan2021}.


\begin{assumption}
 (Value function uniqueness) There exists a unique solution to $v$ in equation~\eqref{eq:boe_almdp} up to a constant shift.
  \label{ass:uniqueness}
\end{assumption}
\begin{assumption} (Stepsize assumption)
    \begin{equation*}
        \alpha_t > 0,\;\sum_{t=0}^{\infty} \alpha_t = \infty,\;\sum_{t=0}^{\infty} \alpha_t^2 < \infty.
    \end{equation*}
      \label{ass:stepsize1}
\end{assumption}
\begin{assumption}
    (Asynchronous Stepsize 1) Let $[\cdot]$ denote the integer part of $(\cdot)$, for $x\in(0, 1)$
    \begin{equation*}
        \sup_i \frac{\alpha_{[xi]}}{\alpha_i} < \infty
    \end{equation*}
    and
    \begin{equation*}
        \frac{\sum_{j=0}^{[yi]}\alpha_j}{\sum_{j=0}^i\alpha_j}\rightarrow 1
    \end{equation*}
    \label{ass:stepsize2}
    uniformly in $y \in[x, 1]$.
\end{assumption}

\begin{assumption}
    (Asynchronous Stepsize 2) There exists $\Delta>0$ such that
    \begin{equation*}
        {\lim\inf}_{t\rightarrow\infty} \frac{\nu(t, s)}{t+1}\geq \Delta
    \end{equation*}
    almost surely, for all $s\in\cS$. Here $\nu(t, s)$ represents the visitation count for state $s$ up to timestep $t$. Furthermore, for all $x > 0$, let
    \begin{equation*}
        N(t, x) = \min \Big\{m > t: \sum_{i=t+1}^m \alpha_i \geq x \Big\}
    \end{equation*}
    \label{ass:stepsize3}
    the limit 
    \begin{equation*}
       \lim_{t\rightarrow\infty} \frac{\sum_{i=\nu(t, s)}^{\nu(N(t, x), s)} \alpha_i}{\sum_{i=\nu(t, s')}^{\nu(N(t, x), s')} \alpha_i}
    \end{equation*}
    exists for all $s, s'\in\cS$.
\end{assumption}

Under the communication assumption, the system
\begin{align}
    v(s) &= \cR(s) - \rho + \frac{1}{\eta} \log \sum {\cP(s'\lvert s) e^{\eta v(s')}},\;\;\forall s\in\cS, \\
    \rho - \hat\rho_0 &= \lambda \Big(\sum_s v(s) - \sum_s\hat v_0(s)\Big),
\end{align}
has a unique solution for $v$ which we denote as $v_\infty$, where $\rho$ is the optimal gain.

At each timestep the increment to $\hat\rho_t$ is $\lambda$ times the increment to $\hat v_t$, and thus, to $\sum_s \hat v_t(s)$. The cumulative increment at $t$ can be expressed as
\begin{align}
     \hat\rho_t - \hat\rho_0 &= \lambda \sum_{i=0}^{t-1}\sum_s \alpha_i(s)\delta_i(s) \indicator{s\in X_t}\nonumber\\
                   &= \lambda \Big(\sum_s\hat v_t(s) - \sum_s\hat v_0 (s)\Big)\nonumber \\
    \implies \hat\rho_t &= \lambda \sum_s \hat v_t(s) - \lambda\sum_s \hat v_0 (s) + \hat\rho_0 \\
    &= \lambda\sum_s \hat v_t(s) - c, \label{eq:cum_rho}\\
    \text{where}\;c &= \lambda \sum_s \hat v_0(s) - \hat\rho_0.
\end{align}

If we replace~\ref{eq:cum_rho} in~\ref{eq:updatev}, we obtain
\begin{equation}
    \widehat{v}_{t+1}(s) \gets \widehat{v}_t(s) + \alpha_t(s) \widetilde\delta_t(s)  \mathbb{I}\{s\in X_t\},~~\forall{s\in\cS},
    \label{eq:td_asynchronous_full}
\end{equation}
where
\begin{equation}
    \widetilde\delta_t(s) = r_t(s) + c - \lambda\sum_s \hat v_t(s) - \frac 1 \eta \log \sum_{s'\in\cS} \cP(s'|s) e^{\eta \widehat{v}_t(s')} - \widehat{v}_t(s).
\end{equation}

This can be interpreted as the TD error of an alternative LMDP $\widetilde\cL=\langle\cS,\cP,\widetilde\cR\rangle$ in which the reward is defined as $\widetilde\cR(s) = \cR(s) + c$ and the gain estimate equals $\lambda \sum_{s\in\cS} \widehat{v}_t(s)$.
The gain of $\widetilde\cL$ satisfies 
\begin{equation}
    \widetilde\rho = \rho + c.
    \label{eq:alternative_reward}
\end{equation}

The former expression, combined with~\eqref{eq:cum_rho} gives
\begin{equation}
    \widetilde\rho = \lambda \sum_s v_\infty.
    \label{eq:extended_rho}
\end{equation}
It is easy to verify that $v_\infty$ is not only the solution for the original LMDP $\cL$, but also for the alternative LMDP $\widetilde\cL$,
\begin{align*}
     v_\infty(s) &= \cR(s) - \rho + \frac{1}{\eta} \log \sum_{s'} {\cP(s'\lvert s) e^{\eta v_\infty(s')}}\;\;\forall s\in\cS\\
     &= \cR(s) - \widetilde\rho + c + \frac{1}{\eta} \log \sum_{s'} {\cP(s'\lvert s) e^{\eta v_\infty(s')}}\;\;\forall s\in\cS\;(\text{by}~\eqref{eq:alternative_reward})  \\
      &= \widetilde\cR(s) - \widetilde\rho + \frac{1}{\eta} \log \sum_{s'} {\cP(s'\lvert s) e^{\eta v_\infty(s')}}\;\;\forall s\in\cS.
\end{align*}

Now consider $\hat\rho_t$. If we can prove that $\hat v_t\rightarrow v_\infty$ then by~\eqref{eq:cum_rho} we have $\hat\rho_t\rightarrow\lambda\sum v_\infty - c$. By~\eqref{eq:extended_rho}, we know that $\lambda\sum v_\infty = \widetilde\rho$, then we have $\hat\rho_t\rightarrow\widetilde\rho-c$. Using~\eqref{eq:alternative_reward}, we get 
\begin{equation*}
    \hat\rho_t\rightarrow\rho\;\text{almost surely as}\;t\rightarrow\infty.
\end{equation*}

The idea is to prove the convergence of differential soft TD-learning for the alternative LMDP $\widetilde\cL$, which is the same solution as for the original LMDP $\cL$.

We adapt Theorem B.2 in~\cite{Wan2021}.

\begin{theorem} (Convergence of differential TD learning)
    For any $v_0\in\real^{\lvert\cS\rvert}$, let $r_t$, $X_t$, $\alpha_t$ be properly defined and consider the update rule
    \begin{equation}
        \widehat{v}_{t+1}(s) \gets \widehat{v}_t(s) + \alpha_t(s) \big( r_t(s) - \lambda\sum_s \hat v_t(s) - \frac 1 \eta \log \sum_{s'\in\cS} \cP(s'|s) e^{\eta \widehat{v}_t(s')} - \widehat{v}_t(s)\big)  \mathbb{I}\{s\in X_t\},
        \label{eq:td_update_theo}
    \end{equation}
    
    \begin{enumerate}
        \item Assumptions~\ref{ass:unichain} and~\ref{ass:uniqueness}-\ref{ass:stepsize3} hold.
        \item $f:\real^{\lvert\cS\rvert}\rightarrow\real$ is Lipschitz and there exists some $u>0$ such that $\forall c\in\real$ and $x\in\real^{\lvert\cS\rvert}$, $f(\allones)=u$, $f(x + c\allones)=f(x)+cu$ and $f(cx) = c f(x)$
    \end{enumerate}
    then $\hat v_t$ converges almost surely to $v_\infty$.
    \label{theo:our_theorem}
\end{theorem}

We observe that~\eqref{eq:td_update_theo} is in the same form of equationB.24 in~\cite{Wan2021} and equation7.1 in~\cite{Borkar2009}. Thus the results in Section 7.4 in~\cite{Borkar2009} and Theorem 3.2~\cite{Borkar1998} apply to show convergence of~\eqref{eq:td_update_theo}. Due to Assumptions~\ref{ass:stepsize2} and ~\ref{ass:stepsize3}, to show convergence of \eqref{eq:td_asynchronous_full} suffices to show convergence of the following synchronous update
\begin{equation}
    \widehat{v}_{t+1} \gets \widehat{v}_t + \alpha_t \pa{ T(\widehat{v}_t) - f(\widehat{v}_t) \mathbbm{1} - \widehat{v}_t + M_{t+1}}.
    \label{eq:synchronous}
\end{equation}

Here, $\hat v_t\in\real^{\lvert\cS\rvert}$ is interpreted as a vector, and the operator $T$ is a mapping $T:\real^d\rightarrow\real^d$ defined for each state $s$ as
\begin{equation*}
    T(v)(s) = \cR(s) + \frac 1 \eta \log \sum_{s'\in\cS} \cP(s'|s) e^{ \eta v(s') }.
\end{equation*}
We also define the following operators, 
\begin{align*}
    T^1(v) &= T(v) - \rho\allones \\
    T^2(v) &= T(v) - f(v)\allones =  T^1(v) + (\rho - f(v))\allones.
\end{align*}

The function $f$ is given by $f(v) = \lambda\sum_s v(s)$, which satisfies condition 2 of Theorem~\ref{theo:our_theorem}. The error term $M_{t+1}=r_t-\cR$ is only needed in case the reward vector $r_t$ is sampled from a distribution with mean $\cR$; if reward is deterministic, $M_{t+1}$ can be omitted. If the reward is sampled from a distribution, it should be easy to show that $M_{t+1}$ has zero mean and bounded variance.

The operator $T$ is a non-expansion in the max-norm:
\begin{align*}
T(x)(s) - T(y)(s) &= \frac 1 \eta \log \sum_{s'} \cP(s'|s) e^{\eta x(s')} - \frac 1 \eta \log \sum_{s'} \cP(s'|s) e^{\eta y(s')}\\
 &= \frac 1 \eta \log \frac {\sum_{s'} \cP(s'|s) e^{\eta x(s')}} {\sum_{s'} \cP(s'|s) e^{\eta y(s')}}\\
 &\leq \frac 1 \eta \log \frac {\sum_{s'} \cP(s'|s) e^{\eta \pa{ y(s') + \infnorm{x-y} }}} {\sum_{s'} \cP(s'|s) e^{\eta y(s')}}\\
 &= \frac 1 \eta \log e^{\eta \infnorm{x-y} } \frac {\sum_{s'} \cP(s'|s) e^{\eta y(s')}} {\sum_{s'} \cP(s'|s) e^{\eta y(s')}}\\
 &= \infnorm{x - y}.
\end{align*}
Hence we have
\[
\infnorm{T(x)-T(y)} = \max_s |T(x)(s) - T(y)(s)| \leq \infnorm{x - y}.
\]

We also show the following property of $T$:
\begin{align*}
T(x + c\mathbbm{1})(s) &= \cR(s) + \frac 1 \eta \log \sum_{s'} \cP(s'|s) e^{\eta \pa{ x(s') + c }}\\
 &= \cR(s) + \frac 1 \eta \log e^{\eta c} \sum_{s'} \cP(s'|s) e^{\eta x(s')}\\
 &= \cR(s) + \frac 1 \eta \sum_{s'} \cP(s'|s) e^{\eta x(s')} + c\\
 &= T(x)(s) + c.
\end{align*}
Hence it follows that $T(x + c\mathbbm{1}) = T(x) + c\mathbbm{1}$.

We consider the following ordinary differential equations (ODEs),
\begin{align}
    \dot{y}_t &= T^1(y_t) - y_t \label{eq:ode1} \\
    \dot{x}_t &= T^2(x_t) - x_t \label{eq:ode2}.
\end{align}
Such equations are well-defined since both RHS's are Lipschitz thanks to the properties of $f$ and $T$.

\noindent We complete the proof as a succession of lemmas.

\begin{lemma}
    Let $\bar{y}$ be equilibrium point of the ODE defined in~\eqref{eq:ode1}. Then $\infnorm{y_t - \bar{y}}$ is non-increasing and $y_t\rightarrow y_\infty$ for some equilibrium point of \eqref{eq:ode2}.
\end{lemma}

\begin{proof}
    See Lemma 3.1 in~\cite{Abounadi2001}.
\end{proof} 

\begin{lemma}
    Equation~\eqref{eq:ode2} has a unique equilibrium at $v_\infty$.
\end{lemma}

\begin{proof}
    See Lemma 3.2 in~\cite{Abounadi2001}.
\end{proof}

\begin{lemma}
    Let $x_0=y_0$, then $x_t = y_t + z_t\allones$ satisfies the ODE $\dot{z}_t = -u z_t + (\rho - f(y_t))$.
\end{lemma}
\begin{proof}
    See Lemma 3.3 in~\cite{Abounadi2001}.
\end{proof} 

\begin{lemma}
    $v_\infty$ is the globally asymptotically stable equilibrium for~\eqref{eq:ode2}
\end{lemma}
\begin{proof}
    See Lemma B.4 in~\cite{Wan2021}.
\end{proof}  

\begin{lemma}
    Equation~\eqref{eq:synchronous} converges almost surely $\hat v_t$ to $v_\infty$ as $t\rightarrow\infty$.
\end{lemma} 
\begin{proof}
    See Lemma B.5 in~\cite{Wan2021} and Lemma 3.8 in~\cite{Abounadi2001}.
\end{proof}

Therefore, stability and convergence of equation~\eqref{eq:td_update_theo} is proved.

\end{document}